\documentclass[10pt,twocolumn,letterpaper]{article}
\usepackage{cvpr}
\usepackage{times}
\usepackage{epsfig}
\usepackage{graphicx}
\usepackage{amsmath}
\usepackage{amssymb}
\usepackage{multirow}
\usepackage{float}
\usepackage{authblk}

\usepackage[pagebackref=true,breaklinks=true,letterpaper=true,colorlinks,bookmarks=false]{hyperref}

\newcommand{\tabincell}[2]{\begin{tabular}{@{}#1@{}}#2\end{tabular}}

\cvprfinalcopy 

\def\cvprPaperID{****} 

\ifcvprfinal\fi
\begin{document}

\title{iMiGUE: An Identity-free Video Dataset for Micro-Gesture Understanding and Emotion Analysis}
\author{Xin Liu$^{\dagger}$, Henglin Shi$^\ddagger$, Haoyu Chen$^\ddagger$, Zitong Yu$^\ddagger$, Xiaobai Li$^\ddagger$, Guoying Zhao$^{\ddagger}$\thanks{Corresponding author. This paper is supported by Academy of Finland, KAUTE foundation, and National Natural Science Foundation of China.}\\
\vspace{-0.3cm}
$^\dagger$School of Electrical and Information Engineering, Tianjin University, China\\
$^\ddagger$Center for Machine Vision and Signal Analysis, University of Oulu, Finland\\
{\href{https://github.com/linuxsino/iMiGUE}{https://github.com/linuxsino/iMiGUE}}}


\maketitle
\thispagestyle{empty}

\begin{abstract}
We introduce a new dataset for the emotional artificial intelligence research: {\bf{\emph{i}}}dentity-free video dataset for {\bf{\emph{Mi}}}cro-{\bf{\emph{G}}}esture {\bf{\emph{U}}}nderstanding and {\bf{\emph{E}}}motion analysis (iMiGUE). Different from existing public datasets, iMiGUE focuses on nonverbal body gestures without using any identity information, while the predominant researches of emotion analysis concern sensitive biometric data, like face and speech. Most importantly, iMiGUE focuses on micro-gestures, \textit{i}.\textit{e}., unintentional behaviors driven by inner feelings, which are different from ordinary scope of gestures from other gesture datasets which are mostly intentionally performed for illustrative purposes. Furthermore, iMiGUE is designed to evaluate the ability of models to analyze the emotional states by integrating information of recognized micro-gesture, rather than just recognizing prototypes in the sequences separately (or isolatedly). This is because the real need for emotion AI is to understand the emotional states behind gestures in a holistic way. Moreover, to counter for the challenge of imbalanced sample distribution of this dataset, an unsupervised learning method is proposed to capture latent representations from the micro-gesture sequences themselves. We systematically investigate representative methods on this dataset, and comprehensive experimental results reveal several interesting insights from the iMiGUE, \textit{e}.\textit{g}., micro-gesture-based analysis can promote emotion understanding. We confirm that the new iMiGUE dataset could advance studies of micro-gesture and emotion AI.
\end{abstract}

\vspace{-0.4cm}
\section{Introduction}
\label{sec:int}

Emotional artificial intelligence (emotion AI) is using machine learning methods to enable computers to understand human emotions. It plays a vital role in human-computer interaction since emotions are on all the time, presented in all kinds of human activities, thinking, and decision makings. According to psychological studies, body language is an essential part for understanding human emotions. Every day, we respond to thousands of such nonverbal behaviors including facial expressions, eye movements or gaze, tone of voices, gestures, touches, and the use of space. Body language-based emotion understanding has attracted extensive attention in the communities of computer vision and affective computing, and a considerable number of datasets have been proposed, \textit{e}.\textit{g}., posed facial expressions \cite{kanade2000comprehensive,pantic2005web, yin20063d, valstar2010induced,gross2010multi,zhang2013high}, spontaneous facial behaviors \cite{aung2015automatic,lucey2011painful,bartlett2006fully,mckeown2011semaine,dhall2017individual,kollias2019deep}, micro-expressions \cite{yan2014casme,li2013spontaneous,davison2016samm}, voice/speech \cite{schuller2011avec, schuller2012avec, mckeown2011semaine, ringeval2013introducing}, social signals \cite{joo2015panoptic,joo2019towards}, and multi-modal datasets with facial expressions and physiological signals \cite{soleymani2011multimodal,koelstra2011deap,mckeown2011semaine,ringeval2013introducing}. Although computational methodologies were proposed correspondingly and consecutively to improve the performance on these datasets, there are still significant gaps between current studies and the needs of real applications. Major limitations include:
\begin{figure}[t]
\begin{center}
\includegraphics[width=8.2cm]{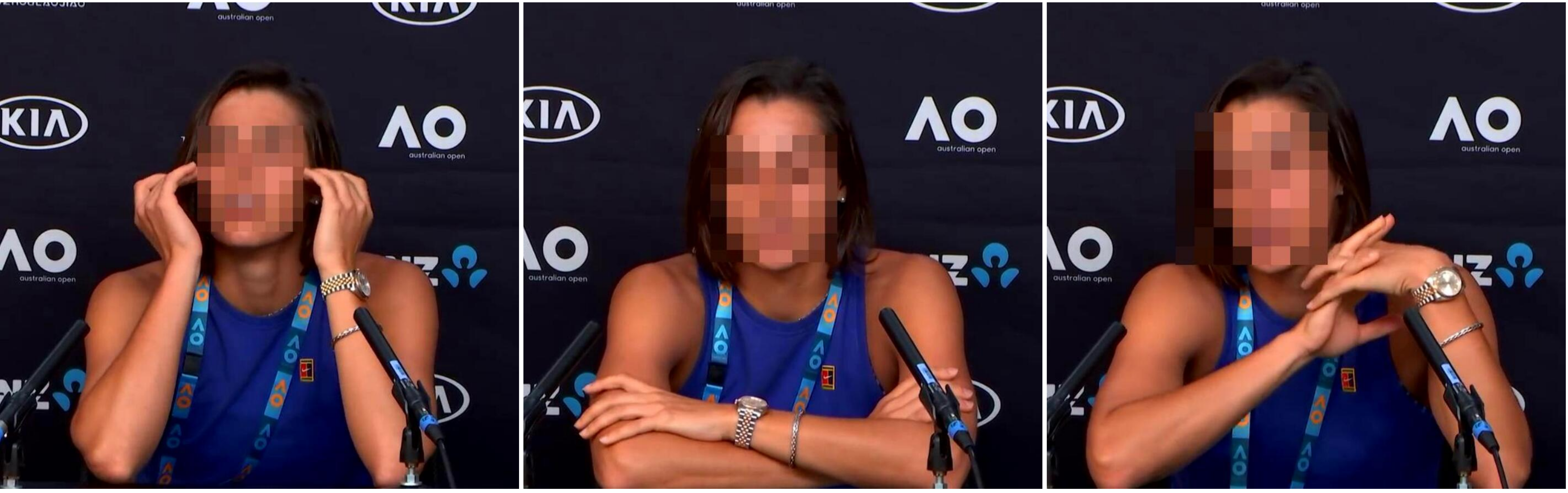}
\end{center}
\vspace{-0.42cm}
\caption{Three frames (cropped) from a post-match press conference video to illustrate the identity-free micro-gestures, such as ``cover face'', ``fold arms'', and ``cross fingers''. Could machine recognize these micro-gestures, and understand emotional states of the player in a holistic way, and further identify if the player has won or lost the match (positive or negative emotional states)?}
\vspace{-0.5cm}
\label{fig:pic1}
\end{figure}
\setlength{\belowcaptionskip}{10pt}

1) {\bf {Intentional behavioral-based gestures}}. Previous gesture studies mostly focused on illustrative (or iconic) gestures \cite{vinciarelli2009social}, \textit{e}.\textit{g}., waving hands as ``hello'' or ``goodbye'', which are intentionally performed for conveying certain meanings or feelings during interactions. However, in many occasions people would suppress or hide their feelings (especially negative ones) rather than expressing them. Previous studies \cite{aviezer2012body,axtell1991gestures,burgoon1989nonverbal} showed that there is a special group of gestures, the micro-gestures (MGs), which are helpful to understand such suppressed or hidden emotions. The major difference between MGs and illustrative gestures is that MGs are unintentional behaviors elicited by people's inner feelings, \textit{e}.\textit{g}., rubbing hands due to stress, and the function of MGs is for relieving or protecting oneself from negative feelings rather than presenting for others. Thus, being able to automatically recognize MGs would allow emotion understanding at a better level. To the best of our knowledge, there is no publicly available dataset for this emotional MGs research in the field of computer vision.

2) {\bf {Gap between behavior recognition and emotion understanding}}. Most existing datasets only aim to evaluate approaches that can detect and recognize prototypes of behaviors (including gestures). In fact, the actual need of emotion AI is not merely to recognize certain behaviors, but to uncover the emotion underneath. Consider a post-match interview scenario, a player is interviewed by reporters over several question \& answer rounds (see Fig.~\ref{fig:pic1}). Some MGs could be observed, \textit{e}.\textit{g}., cross arms (defensive) and cover face (upset or ashamed), but it is hoped that the machine can understand (identify) if the player has a positive or negative feeling (\textit{e}.\textit{g}., caused by winning or losing of the match).

3) {\bf {Sensitive biometric data}}. Most of the existing datasets involve sensitive biometric data. Actually, biometric data based identity recognition plays a critical role in a variety of applications and has gained great success in the past decade. While every coin has two sides, biometric information is so sensitive that is particularly prone to be (identity) stolen, misused, and unauthorized tracked. With the concerns of privacy grows, more attention should be paid to protect biometric data of individuals.

Psychological studies \cite{ekman2009telling} showed that there are over 215 behaviors associated with psychological discomfort and most of them are not in the face. MGs are subtle and (some of them) short, mostly out of our awareness or notice during live interactions \cite{ginger2007gestalt}. It would be of great value if we can develop computer vision methods to capture and recognize these neglected clues for better emotion understanding. In this paper, we introduce a novel MGs dataset to address afore mentioned limitations. The key contributions are summed up as follows:

1) Instead of using facial or vocal-expressions for emotion understanding, the proposed dataset offers an approach where the identity-free MGs are explored for hidden emotion understanding, and privacy of the individuals could be preserved. As far as we know, iMiGUE is the first public benchmark focuses on emotional MGs. This is the first investigation of such gestures from the computer vision perspective. Moreover, to deal with the issue of imbalanced classes distribution, an unsupervised model is proposed.

2) iMiGUE is not only for MG recognition, but also provides a hierarchy that allows exploration of the relationship between MGs and emotion, \textit{i}.\textit{e}, associates the MGs holistically for emotion understanding. As such, the data in iMiGUE are annotated on two levels: the MG categories were annotated on video clip-level, and the emotion categories were labeled on video-level.

3) Comprehensive experiments are conducted on the iMiGUE to provide baseline results. In video clip-level, the experimental results show that even fully supervised learning SOTA models cannot yield satisfactory accuracy on iMiGUE, which could verify that the challenges of recognizing such hardly noticeable MGs. The proposed unsupervised method can achieve competitive performance compared with many supervised models. In video-level, we find micro-gesture is a vital factor for emotion understanding. We only employed a simple recurrent neural network (RNN) network to achieve MGs analysis in a holistic way, but its emotion understanding result can beat existing action/gesture recognition-based models. The dataset and findings will serve as a launch-pad for exploring identity-free MG-based emotion AI.

\section{Related Work}
A person's emotional state is often conveyed through bodily expression. As such, analyzing body based activities, including action, gesture and posture are the popular research topics \cite{soomro2012ucf101,kuehne2011hmdb, gorban2015thumos,rohrbach2016recognizing,shahroudy2016ntu,sigurdsson2016hollywood, goyal2017something,monfort2019moments,yu2020humbi} in the community. However, these datasets focused on recognizing human activities (\textit{e}.\textit{g}., a man is jumping), rarely related to the emotional states. We limit our review on the related emotional gesture-based benchmarks. Then we review related work of gesture/action recognition.
\begin{table*}
  \centering
  \footnotesize
    \begin{tabular}{|l|c|c|c|c|c|c|c|c|c|c|c|}
    \hline
     \tabincell{c}{Datasets} & \tabincell{c}{\texttt{\#} Ge-\\stures} & \tabincell{c}{\texttt{\#} Em-\\otions}  &  \tabincell{c}{\texttt{\#}Subjects\\(F/M)}  & \tabincell{c}{\texttt{\#} Sam-\\ples} & \tabincell{c}{\texttt{\#}Vid-\\eos} & \tabincell{c}{Duration}  & \tabincell{c}{Con-\\text} & \tabincell{c}{Expr-\\ession} & \tabincell{c}{Resolution} & \tabincell{c}{Modalities} & \tabincell{c}{Recog-\\nition}\\
    \hline \hline
FABO \cite{gunes2006bimodal}                  & - & 10 & 23  (12/11)  & 206   & 23   & 6 Min      & C  & Posed & 1024$\times$768   &F + G  &Isolated\\
HUMAINE \cite{douglas2007humaine}             & 8 &  8 & 10  (4/6)    & 240   & 240  & 5-180 Sec  & C  & Posed & -                &F + G  &Isolated\\
GEMEP \cite{glowinski2008technique}           & - & 18 & 10 (5/5)  & 7\,000+  & 1260 & -        & C  & Posed & 720$\times$576    &F + G  &Isolated\\
THEATER \cite{kipp2009gesture}                & - & 8  & -            & 258   & -    & -        & U  & SP  & -                &G  &Isolated\\
EMILYA \cite{fourati2014emilya}               & 7 & 8  & 11 (6/5)   & 7\,084  & 23   & 5.5 Sec    & C  & Posed &1280$\times$720   &G  &Isolated\\
LIRIS-ACCEDE \cite{gavrilescu2015recognizing} & 6 & 6  & 64  (32/32)  & -     & -    & 1 Min      & C  & Posed & -                 &F + G  &Isolated\\
emoFBVP \cite{ranganathan2016multimodal}      & 23 & 23  & 10 (-)  & 1\,380  & -    & 20-66 Sec   & C  & Posed &640$\times$480   &F + G + V  &Isolated\\
BoLD \cite{luo2020arbee}                      & - & 26 & -    & 13\,239  & 9\,876    & -        & U  & SP  & -   &G  &Isolated\\
    \hline
iMiGUE (Ours)                                 & 32 & 2 & 72 (36/36)   & 18\,499 & 359 & 0.5-25.8 Min & U  & SP   & 1280$\times$720 & IMG     & Holistic\\
    \hline
    \end{tabular}
    \setlength{\abovecaptionskip}{0pt}
      \caption{The attributes comparison of iMiGUE with other widely used datasets for recognizing gesture-based expression of emotion. F/M: Female/Male, C: Controlled (in-the-lab), U: Uncontrolled (in-the-wild), SP: Spontaneous, F: Face,  G: Gesture, V: Voice, IMG: Identity-free Micro-Gesture.}
      \vspace{-0.4cm}
\label{tb:datasetscom}
\end{table*}%

\subsection{Emotional Gesture-based Datasets}
Gesture is one of the key cues of social communication which includes movements of hands, head and other parts of human body that express various feelings, thoughts and emotions \cite{noroozi2018survey}. Table \ref{tb:datasetscom} summarizes the attributes of widely used databases of emotional gestures. In this field, early studies were mostly built on acted or posed gestures. The Tilburg University Stimulus set \cite{schindler2008recognizing} collected photographic still images of 50 actors enacting different emotions. FABO database \cite{gunes2006bimodal} is one of the pioneer work which proposed using video clips of posed prototype gestures to recognize emotions. These videos were labelled with six basic emotions, as well as four more states, namely, neutral, anxiety, boredom, and uncertainty. Following that posed behavior which was captured in controlled recording conditions, researchers extended emotional gesture analysis into many directions. In HUMAINE \cite{douglas2007humaine,castellano2007recognising}, the researchers elicited emotions via interaction with computer avatar of its operator. The Geneva multi-modal emotion portrayals (GEMEP) database \cite{glowinski2008technique} contains more than 7\,000 audio-video portrayals of 18 emotions portrayed by 10 actors. Also in a controlled setting, the subset \cite{gavrilescu2015recognizing} of LIRIS-ACCEDE database \cite{baveye2015liris} collected upper body emotional gestures from 64 subjects. Using a Kinect sensor, Saha \textit{et al}. \cite{saha2014study} collected 3D skeleton gesture data of 10 subjects, which included five induced emotions, \textit{i}.\textit{e}., anger, fear, happiness, sadness, and relaxation. Psaltis \textit{et al}. \cite{psaltis2016multimodal} collected skeletal gestural expressions that frequently appear in a game-play scenario. Similarly, the emoFBVP \cite{ranganathan2016multimodal} dataset has a multi-modal recordings of actors including body gestures with skeletal tracking. Emilya \cite{fourati2014emilya} dataset captured 3D body movements of posed emotions via a motion capture system.

Later studies focused more toward spontaneous emotional gestures, which are more challenging than posed ones. In the Theater \cite{kipp2009gesture} dataset, the emotional gesture video clips were extracted from two movies, which are close to real world scenes. In \cite{kleinsmith2011automatic}, gesture movements are recorded while subjects were playing body movement based video games. Luo \textit{et al}. collected a large-scale bodily expression dataset, the BoLD \cite{luo2020arbee}, in which the in-the-wild perceived emotion data were segmented from movies and reality TV shows. To date, few efforts were made on the fine-grained micro visual of the body, \textit{i}.\textit{e}., the MG, which is important clue for understanding suppressed or concealed emotions.

\subsection{Methods for Gesture/Action Recognition}
Early work of automatic modeling of emotional gestures depends largely on hand-crafted features \cite{gunes2006bimodal,schindler2008recognizing,kleinsmith2011automatic}. Recently, numerous neural networks have been introduced for gesture/action recognition. Among them, supervised learning is the predominate technique for which labeled data are utilized to train the models. The earliest attempts utilized a 2D convolutional neural network (CNN) \cite{simonyan2014two, feichtenhofer2016convolutional,wang2018temporal,zhou2018temporal,lin2019tsm,yu2020searching} to extract spatial features from the selected frames, and the temporal aggregation is considered by an additional stream of optical flow or the temporal pooling layers. The 3D CNNs \cite{tran2015learning,carreira2017quo,wang2018non,tran2018closer,hara2018can,xie2018rethinking,tran2019video} can jointly capture spatial-temporal semantics, where the filters are designed in a 3D manner. Compared to the 2D CNNs, 3D ones can process the temporal information hierarchically throughout the whole network. Also, some models like the Slow-fast \cite{feichtenhofer2019slowfast} considered a joint implementation of both two streams (fast and slow) and 3D CNN. The RNN is also commonly used for temporal integration. Specifically, the long short-term memories (LSTMs) \cite{donahue2015long, du2015hierarchical, veeriah2015differential, zhu2016co,Mahasseni_regularized_lstm:2016, li2016online,liu2016spatio,liu2017global} have demonstrated their strength on learning sequential data. Recently, the skeleton data is gaining increasingly popularity because of their invariance to background dynamics. Current work on skeleton based action recognition can mainly be categorized into two types: one is RNN based methods \cite{du2015hierarchical,liu2016spatio,song2017end,liu2017global,zhang2017view,li2018independently} which directly process gesture skeletons as time series, and the other one is graph convolutional network (GCN) \cite{yan2018spatial,li2018spatio,si2019attention,shi2019two,peng2020learning,cheng2020skeleton, liu2020disentangling} based methods which reorganize the skeleton data as a graph.

\begin{figure*}[t]
\begin{center}
\includegraphics[width=14cm]{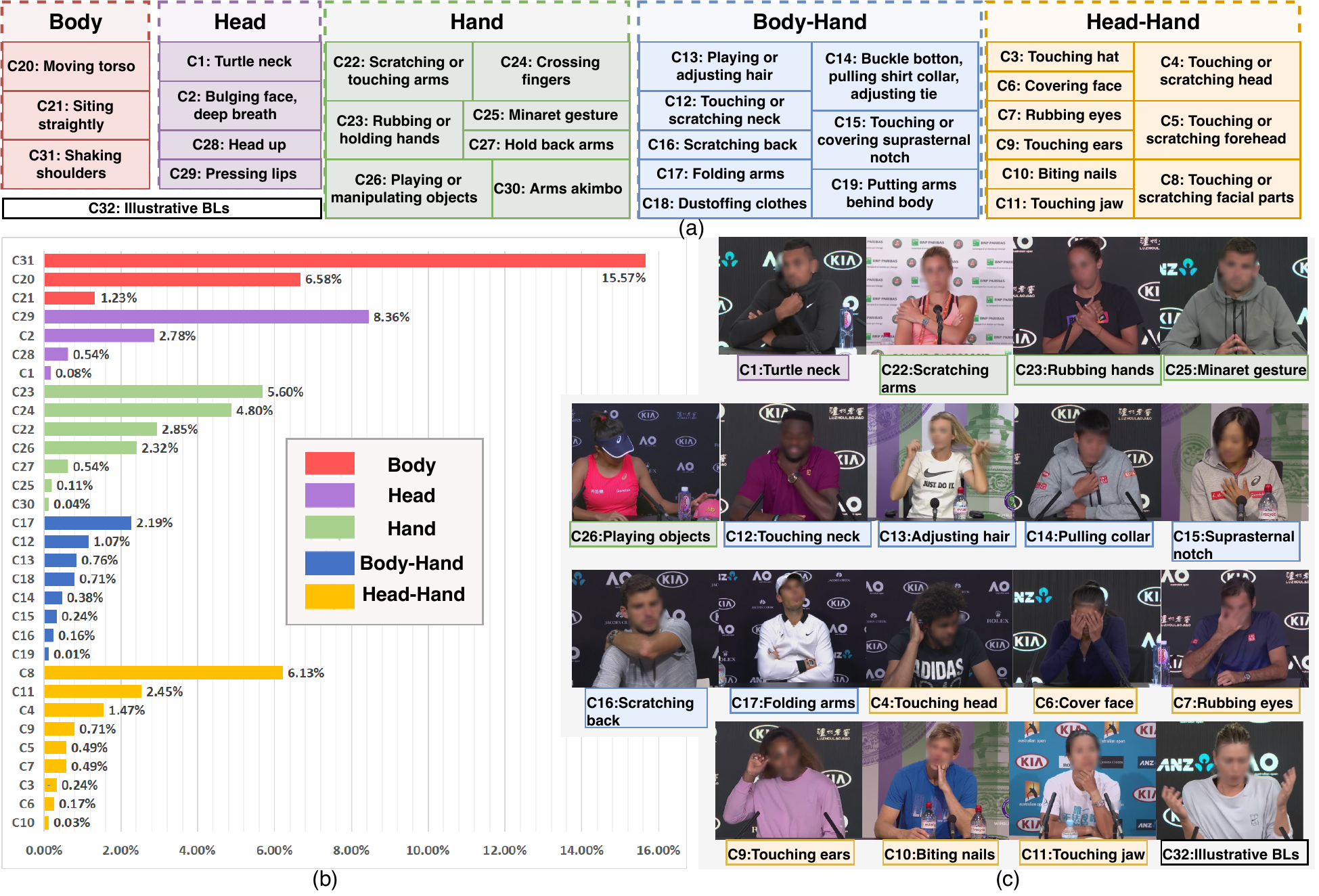}
\end{center}
\vspace{-0.45cm}
\caption{(a) Categories of MGs in iMiGUE dateset which refers to psychological studies \cite{ekman2009telling,pease2008definitive,navarro2016every}. (b) Sample percentages of each category in iMiGUE dataset. (c) Examples (face masked) of micro-gesture categories in iMiGUE dataset.}
\vspace{-0.4cm}
\label{fig:imigue}
\end{figure*}

Compared with the supervised methods, the task of behavior recognition with unsupervised approaches is much more challenging, and only a few attempts have been reported. Some methods focused on leveraging temporal information of videos to learn visual representation, such as Shuffle \& Learn \cite{misra2016shuffle}, OPN \cite{lee2017unsupervised}, and \cite{xu2019self,tao2020self,wang2019self,kim2019self}. Other methods utilized the encoder-decoder-based video sequence/frame reconstruction, \textit{e}.\textit{g}., RGB-based \cite{srivastava2015unsupervised,luo2017unsupervised,li2018unsupervised}, and skeleton-based LongT GAN \cite{zheng2018unsupervised} and Predict \& Cluster (P\&C) \cite{su2020predict}. The problem is that P\&C just used the reconstruction loss in an element-wise manner without considering any informative constraint or prior. In addition, P\&C employed a fixed-length input scheme which is hard to encode the long-term motion dependencies since the down-sampled sequences may lose essential information.

\section{The iMiGUE dataset}
\subsection{Key Challenges}
We hope to draw more attention on analyzing micro-gestures starting with building and sharing a new MG database. We need to solve several unprecedent challenges to build the iMiGUE as it is different than previous gesture dataset. 1) \noindent{\bf{\emph{How to define and organize the categories of MGs related to emotions?}}} We take reference from psychological studies \cite{ekman2009telling,pease2008definitive,navarro2016every} when considering the categories of MGs, the focus is to clarify the boundary to differentiate MGs from illustrative behaviors. 2) \noindent{\bf{\emph{How to elicit or collect the spontaneous MGs?}}} The new dataset will definitely contain genuine expressions but not posed ones, and we think it is good to start with selecting and collecting real world videos that contain MG occasion, \textit{e}.\textit{g}., from the online video-sharing platforms. Furthermore, as an important goal of iMiGUE is to study the emotional states behind MGs, we need to find and provide the root cause of these (MGs) emotional outbursts. 3) \noindent{\bf{\emph{How to annotate the data?}}} The quality of labeling will greatly influence the final results of model training and testing. We need to annotate the data on two levels, first is the labels of all MG occasions (clip-level), and second is the labels of corresponding emotion (video-level). The MG labels are based on criteria from related psychological studies. We consider ``positive'' and ``negative'' as two emotional categories to start with, and the labels are based on objective facts, \textit{i}.\textit{e}., winning (positive emotion) or losing (negative emotion) of a match. Considering the complexity and diversity of the MGs, a team trained specifically for the annotation job has been hired. To further make sure the high-quality annotation, an efficient mechanism for quality control has been designed (see Sec. \ref{sec:datacon}).

\subsection{Dataset Construction}
\label{sec:datacon}
\noindent{\bf{Data collection.}} Based on the above factors, we search and collect videos containing scenarios of ``post-match press conferences'' (see Fig. \ref{fig:pic1}), in which a professional athlete was interviewed by journalists and reporters over several question \& answer rounds after a tough match. The players had no (or a little) time to prepare as the press conference will be held right after the match, and he or she needs to respond to the questions rapidly. Even though an experienced person could respond with ``witty'' statements, the emotion-related MGs may leak out unintentionally and unconsciously, since ``the subconscious mind acts automatically and independently of our verbal lie'' \cite{pease2008definitive}. The result of the match, winning or losing is a natural emotion inducer leads to positive or negative emotion states of the interviewed player. Accordingly, for the research purpose of emotion AI, these identity-free MGs should be recognized and understood by machines, and a real AI could use them holistically to understand emotional state of the player.

We choose the post-match press conferences of ``Grand Slam'' (tennis) tournaments as the first data-source for iMiGUE, because they have several merits: 1) There are large numbers of openly accessed videos with high recording quality, \textit{e}.\textit{g}., resolutions of at least 720P, so that even subtle differences between MG instances are well preserved. 2) MG-centric instances without background interference. Unlike several datasets where the background is either disturbing or distinguishable for different categories, all post-match press conference videos were recorded with the same static advertising wall background. 3) Diversity of cultures and nationalities. The players of tournaments come from almost every corner of the Earth (see statistics of iMiGUE in Sec.\ref{sec:datastat}). 4) Good gender balance. Each Grand Slam tournament has 128 male and 128 female players, so it is easy to setup a gender-balanced dataset.

\noindent{\bf{Data Annotation.}}
The collected videos are annotated on two levels: the MG categories, and the emotion categories. The emotions are annotated on video level, \textit{i}.\textit{e}., one emotion label for each press conference video. We consider two emotion categories: positive, \textit{i}.\textit{e}., corresponding to the winning case, and negative, \textit{i}.\textit{e}., corresponding to the losing cases. Then we search through the videos to spot and exert all MG instances (clips) and assign MG category labels for them. The work of MG annotation was very difficult and time consuming, and we took three measures as follows to ensure the quality of annotation.
1) {\bf{\emph{Clarify the scope and categories of MGs.}}}
According to reference psychological studies \cite{ekman2009telling,pease2008definitive,navarro2016every}, MGs could be clusters as five major groups according to the motions' locations and functions, \textit{i}.\textit{e}., ``Head'', ``Body'', ``Hand'', ``Body-Hand'', and ``Head-Hand'', and each major group contains multiple fine-categories of MGs. The iMiGUE covers altogether 31 categories of MGs plus one extra category of non-MGs, \textit{i}.\textit{e}., illustrative gestures  (see Fig. \ref{fig:imigue} (a) for details).
2) {\bf{\emph{Multiple labelers and training for labeling.}}}
We have five persons working together on MG annotation for two reasons: the first is to speed up the process, and the second is to reduce personal bias for more reliable annotation. Before the actual annotation, all five were trained to unify their criteria for MG annotation. First, they went through the descriptions and sample figure or video of the 32 categories of iMiGUE to get understanding of the characteristics of MGs, and primarily rules of instance durations (the starting and ending points) were also discussed. Then, they went through three rounds of labeling exercises, \textit{i}.\textit{e}., in each exercise, every labeler first labeled two sample clips separately and then compared their labels together, different opinions were carefully discussed until agreements were reached by all annotators.
3) {\bf{\emph{Cross check for reliable annotations.}}}
The task of annotating all video clips was divided for five persons to ensure that every clip has two labelers. After all five labelers finished labeling, a cross check of their annotations were carried out following the Eq. \ref{eq:reliab}
\begin{equation}\label{eq:reliab}
\mathcal{R} = \frac{2\times MG(L_{i},L_{j})}{\# All\_MG},
\end{equation}
where $MG(L_{i},L_{j})$ is the number of MGs on which Labeler i and Labeler j agreed, and $\# All\_MG$ is the total number of MGs annotated by the two labelers. The average inter-labeler reliability $\mathcal{R}_{avg}$ of iMiGUE is 0.81 which indicates reliable annotations. For the inconsistent annotation cases, the five labelers discussed them through and kept those with unified opinions while the rest (still with diverse opinions) were left out of the final label list.

\subsection{Dataset Statistics and Properties}
\label{sec:datastat}
iMiGUE collected 359 videos (258 wins and 101 losses) of post match press conferences of Grand Slam tournaments from online video sharing platforms, \textit{e}.\textit{g}., YouTube, of the total length of 2\,092 minutes. The videos' duration varies with an average length of 350 seconds. The videos' resolution is 1280$\times$720, and their frame rate is 25 fps. A total of 18\,499 MG samples were labeled out and assigned with 32 category labels, \textit{i}.\textit{e}., about 51 MG samples each video on average. The length of MG instances also varies, from 0.18s (second) to 80.92s with an average duration of 2.55s. Table \ref{tb:datasetscom} shows the key characteristics numbers of the iMiGUE compares with other gesture datasets. Notice that the sample numbers of the 32 MG categories vary a lot (see Fig. \ref{fig:imigue} (b)), which is a common situation in many spontaneous emotion datasets \cite{yan2014casme,li2013spontaneous,davison2016samm} as it is not control-recorded data and the occurrence of different behaviors naturally varies. The sample unbalance is one challenge for MG recognition, which we will elaborate later in Sec. \ref{sec:unsup}.

This iMiGUE dataset has some attracting properties that distinguishes it from existing work. 1) {\bf{Micro-gesture-based dataset.}} To the best of our knowledge, this is the first public dataset of micro-gestures, which is built to analyse these very fine clues with computer vision methods for recognizing and understanding suppressed or concealed emotions. 2) {\bf{Identity-free.}} The sensitive biometric data, such as the face and voice have been masked and removed. 3) {\bf{Ethnic diversity.}} iMiGUE contains 72 players from 28 countries and regions (\textit{e}.\textit{g}., Argentina, Australia, Spain, Canada, China, United States, and South Africa) covering every continent which enables MGs analysis from diverse cultures. 4) {\bf{Gender-balanced.}} iMiGUE comprises 36 female and 36 male players whose ages are between 17 and 38. 5) {\bf{Winning and losing as the natural and objective reference for emotion categories.}} The iMiGUE is built not only for MG recognition but more importantly for exploring the relationship of MGs and the emotional states. As a new dataset with many unestablished factors, instead of arbitrarily assigned emotion labels which could be biased by subjective judgments, the results of matches could serve as a more objective reference of emotional states, \textit{i}.\textit{e}., to assume that winning a match would lead to a more positive emotion status than losing one. Because this dataset is to analyze MGs and further recognize suppressed emotions without using sensitive biometric data, we suggest the researchers who work on estimating emotional states but concern the privacy issues could use this dataset as a benchmark.

\begin{figure}[t]
\begin{center}
		\centering
		\includegraphics[scale=0.34]{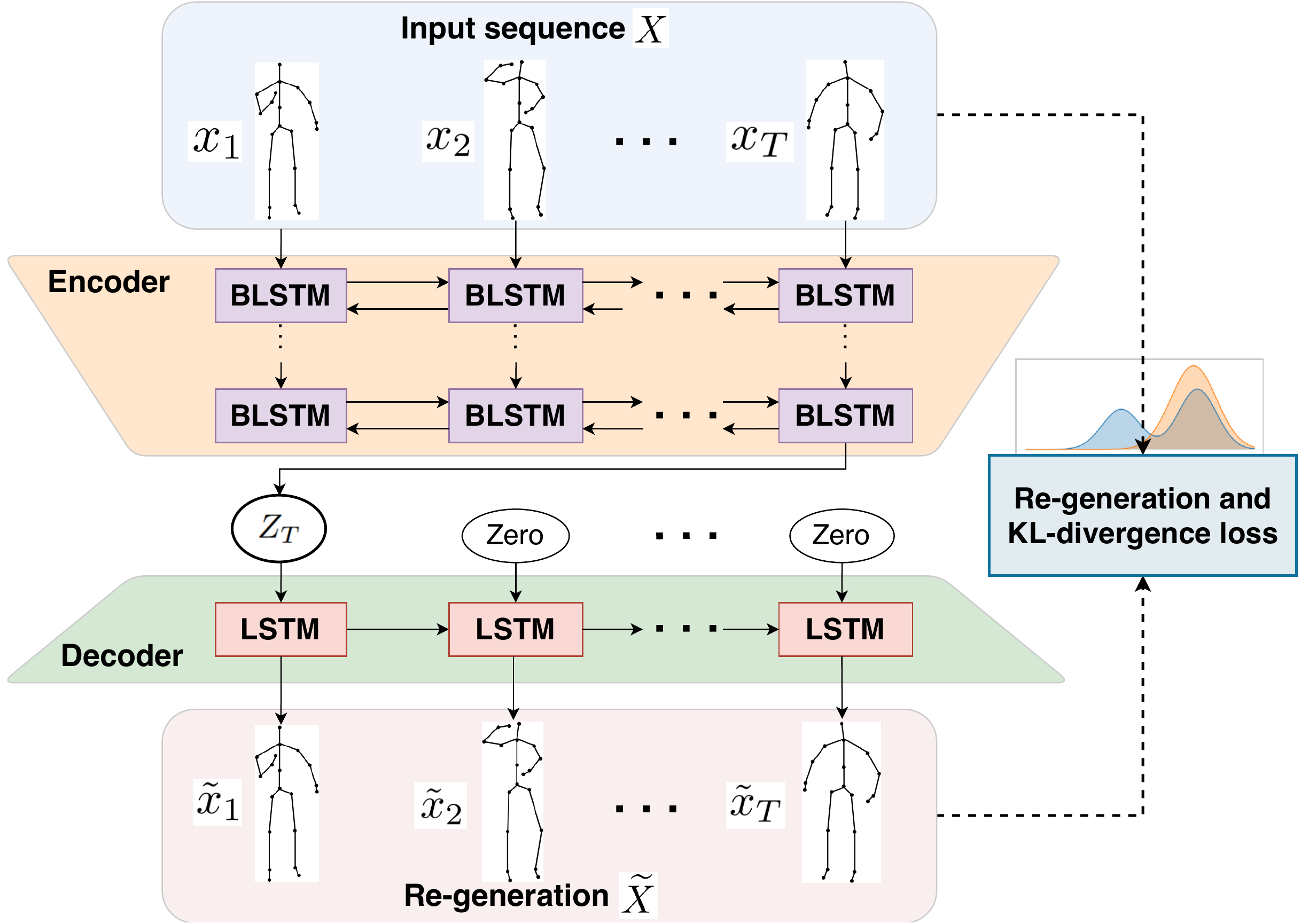}
\end{center}
\vspace{-0.42cm}
\caption{Framework of the proposed unsupervised encoder-decoder network.}
\vspace{-0.3cm}
\label{fig:pip}
\end{figure}

\section{Unsupervised Learning for MG Recognition}
\label{sec:unsup}
After the dataset building, a challenging issue is discovered and should be carefully discussed. Compared with the controlled recording circumstance with the fixed or planed number of samples, the imbalanced data issue is hard to avoid in the condition of an in-the-wild setting. In other words, iMiGUE dataset has a long-tailed distribution with class imbalance issue (see Fig. \ref{fig:imigue} (b)), which may raise a challenge for fully-supervised learning models and cause a significant performance drop under the extreme label bias.

As an intuitive substitution of a fully supervised method, the unsupervised model is advantageous since it does not require human-labeled data. In this paper, following the Seq2Seq unsupervised learning routine \cite{srivastava2015unsupervised,su2020predict}, we introduce an encoder-decoder model to learn discriminative information of MG (pose or key-points-based) sequences without using labelled data. There are a few key differences between our method and previous unsupervised models: 1) We introduce the mutual information to control characteristics of the representation by matching to a prior distribution adversarially, namely, the Kullback-Leibler (KL) divergence is utilized to act as a measure of non-linear statistical dependence between input sequence and reconstructed one which facilitates the model to learn inherent action/gesture representations. While most of existing methods only rely on traditional element-wise loss, \textit{e}.\textit{g}., mean square error ($L_2$ distance) \cite{zheng2018unsupervised} \cite{su2020predict} and mean absolute error ($L_1$) \cite{su2020predict}. 2) Unlike other Seq2Seq type encoder-decoder with a fixed-length scheme \cite{srivastava2015unsupervised} \cite{su2020predict} that only read in parts of the input sequence, we provide a flexible strategy which enables the encoder to read in the whole input sequence, aiming to utilize the complete context information and capture the long-term dynamics in sequences with arbitrary lengthes. To realize the above functions, we extend the preliminary model of sequential variational autoencoder (S-VAE) \cite{shi2018bidirectional}, which is a variant of VAE whose encoder-decoder are implemented by bidirectional LSTM (BLSTM). Different to the vanilla VAE, S-VAE can handle sequential data and capture latent patterns from the whole input sequence.

The framework of the proposed unsupervised S-VAE (U-S-VAE) is illustrated in Fig. \ref{fig:pip}. The encoder of U-S-VAE is a multi-layer BLSTM in which the input is a whole sequence of body key-points (pose) corresponding to an MG $X=(x_1,x_2,....,x_T)$. After the last frame is read in, the hidden state $Z_{T}$ is passed to the decoder which acts as the holistic summary of $X$. During the decoding phase, a simple LSTM decoder receives the $Z_T$ at the first time-step and further re-generates the whole input sequence, denoted as $\widetilde{X}=(\tilde{x}_1,\tilde{x}_2,....,\tilde{x}_T)$. In particular, we train U-S-VAE with a joint loss function:
\vspace{-0.1cm}
\begin{equation}\label{eq:loss}
\mathcal{L}_{\rm{joint}} = \mathcal{L}_{\rm{reg}}+\lambda\mathcal{L}_{\rm{KL}},
\vspace{-0.1cm}
\end{equation}
where $\mathcal{L}_{\rm{reg}}$ is the element-wise-based re-generation loss ($||X-\widetilde{X}||^2_2$) which is responsible for ensuring overall structure similarity between the input and the reconstructed one. Importantly, the KL-divergence is introduced to ensure closer approximation to the joint distribution and the product of the marginals. The motivation is to train a representation-learning encoder-decoder to maximize the mutual information between the inputs and the re-generated. The $\lambda$ controls the weight of KL-divergence loss. As shown in Fig. \ref{fig:pip}, the decoder LSTM reads in the $Z_T$ as the first-frame data to initiate its states. In each of the next time step, instead of the masked ground truth \cite{zheng2018unsupervised} or any other meaningful information as an input, zeros are being fed into the decoder. This operation aims to weaken the decoder which cannot get any information for prediction and it exclusively relies on the state $Z_T$ passed by the encoder. In other words, this strategy enforces the encoder to learn the latent features and represent them with the final state transferred to the decoder. After this unsupervised network is trained, the latent state $Z_T$ of the encoder can be used for classifying MG. Similar to \cite{su2020predict}, for the feature vectors $Z_T$ of all sequences in the training set, a $K$-nearest neighbors (KNN) classifier is used to assign classes.

\section{Experiments}
\label{sec:exp}
\subsection{Benchmark Evaluations}
\label{sec:eva}
To have standard evaluations for all the reported results on the iMiGUE dataset, a two-level criteria has been defined. More specifically, on the MG recognition level, we utilized the cross-subject evaluation protocol which divides the 72 subjects into a training group of 37 subjects and a testing group of 35 subjects. The training and testing sets have 13\,936 and 4\,563 MG samples, respectively. The IDs of training and testing subjects can be found in the supplementary materials. On the emotion classification level, we selected 102 videos (51 win and 51 lose matches) as the training set, and 100 videos (50 win and 50 loss matches) as the test set. The player's emotional states (positive/negative) with the result of win or loss, are classified via analysis of MGs. The details of training and testing protocols (video IDs) can be found in the supplementary material. Specially, to benefit the community of skeleton or pose-based gesture recognition, we provide the pose data of every frame, achieved by using the OpenPose toolbox \cite{cao2017realtime}.

\subsection{Implementation details}
In the proposed U-S-VAE, we set the following architecture: Encoder: 1-Layer BLSTM with N = 256 units for each direction. Decoder: 1-Layer LSTM with N = 256 units. The learning rate is 0.0002 with a decay factor of 0.999 for every five training epochs. The network is trained till the loss converges such that the training loss tends to be stable.

A series of experiments are conducted on the iMiGUE dataset on a PC with a Titan RTX GPU. All training configurations follow the original papers unless stated otherwise.

\begin{table}
\centering
\footnotesize
\begin{tabular}{|c|l|c|c|c|}
\hline
  \multicolumn{2}{|c|}{\multirow{2}{*}{\tabincell{l}{Methods}}} & \multirow{2}{*}{\tabincell{l}{Model+Modality}} & \multicolumn{2}{c|}{Accuracy}\cr\cline{4-5}
  \multicolumn{2}{|c|}{}  &  & Top-1 & Top-5\\
    \hline \hline
 \multirow{15}{*}{\tabincell{l}{Super-\\vised}}     &      S-VAE \cite{shi2018bidirectional}   &   \multirow{3}{*}{\tabincell{l}{RNN + Pose}}        & 27.38 & 60.44\\
      &      LSTM                                &                               & 32.36 & 72.93\\
      &      BLSTM                               &                               & 32.39 & 71.34 \\
    \cline{2-5}
      &      ST-GCN \cite{yan2018spatial}        &  \multirow{5}{*}{\tabincell{l}{GCN + Pose}}         & 46.97 & 84.09\\
      &      2S-GCN \cite{shi2019two}            &                               & 47.78 & 88.43\\
      &      Shift-GCN \cite{cheng2020skeleton}  &                               & 51.51 & 88.18\\
      &      GCN-NAS \cite{peng2020learning}     &                               & 53.90 & 89.21\\
      &      MS-G3D \cite{liu2020disentangling}  &                               & 54.91 & \underline{89.98}\\
    \cline{2-5}
      &      C3D \cite{tran2015learning}	        &	\multirow{3}{*}{\tabincell{l}{3DCNN + RGB}}	 & 20.32  	& 55.31	\\
      &      R3D-101  \cite{hara2018can}		    &		                     & 25.27     & 59.39 \\
      &      I3D  \cite{carreira2017quo}         &                               & 34.96     & 63.69 \\
    \cline{2-5}
      &      TSN \cite{wang2018temporal}         &   \multirow{3}{*}{\tabincell{l}{2DCNN + RGB}}     & 51.54  	& 85.42 \\
      &      TRN \cite{zhou2018temporal}         &                              & \underline{55.24}  	& 89.17	\\
      &      TSM \cite{lin2019tsm}               &                               & \textbf{61.10}  	& \textbf{91.24} \\
    \hline
\multirow{2}{*}{\tabincell{l}{Unsup-\\ervised}}      &      P\&C \cite{su2020predict}                   &   \multirow{2}{*}{\tabincell{l}{Encoder-Decoder\\+ Pose}}             & 31.67 & 64.93\\
      &      U-S-VAE Z (Ours)                            &                                     & 32.43 & 64.30\\
   \hline
\end{tabular}
\setlength{\abovecaptionskip}{0pt}
\caption{Comparison of MG recognition accuracy (\%) with state-of-the-art algorithms on the iMiGUE dataset (best: bold, second best: underlined).}
\vspace{-0.4cm}
\label{tb:mgreg}
\end{table}

\subsection{Clip-level Micro-gesture Recognition}
\label{sec:clip}
In order to evaluate supervised learning-based methods' performance on iMiGUE, 14 state-of-the-art algorithms are selected which can be simply categorized into four groups, namely, body key-points-based RNN (\textit{i}.\textit{e}., BLSTM, LSTM, and S-VAE \cite{shi2018bidirectional}), and GCN (\textit{i}.\textit{e}., ST-GCN \cite{yan2018spatial}, 2S-GCN \cite{shi2019two}, Shift-GCN \cite{cheng2020skeleton}, GCN-NAS \cite{peng2020learning}, and MS-G3D \cite{liu2020disentangling}), RGB-based 3DCNN (\textit{i}.\textit{e}., C3D \cite{tran2015learning}, R3D-101 \cite{hara2018can}, and I3D \cite{carreira2017quo}), and 2DCNN with temporal reasoning (\textit{i}.\textit{e}., TSN \cite{wang2018temporal}, TRN \cite{zhou2018temporal}, and TSM \cite{lin2019tsm}). We further evaluate the effectiveness of the proposed U-S-VAE by comparing it with existing unsupervised methods. In fact, only a few pose (skeleton)-based unsupervised models were proposed, \textit{e}.\textit{g}., LongT GAN \cite{zheng2018unsupervised} and P\&C \cite{su2020predict}. Here, we report the results of P\&C since its implement code is publicly available. It is noted that all models follow the same evaluation protocol mentioned above for a fair comparison. In Table \ref{tb:mgreg}, we present the performances of these baseline networks.

\begin{figure}[t]
\begin{center}
\includegraphics[width=8.3cm]{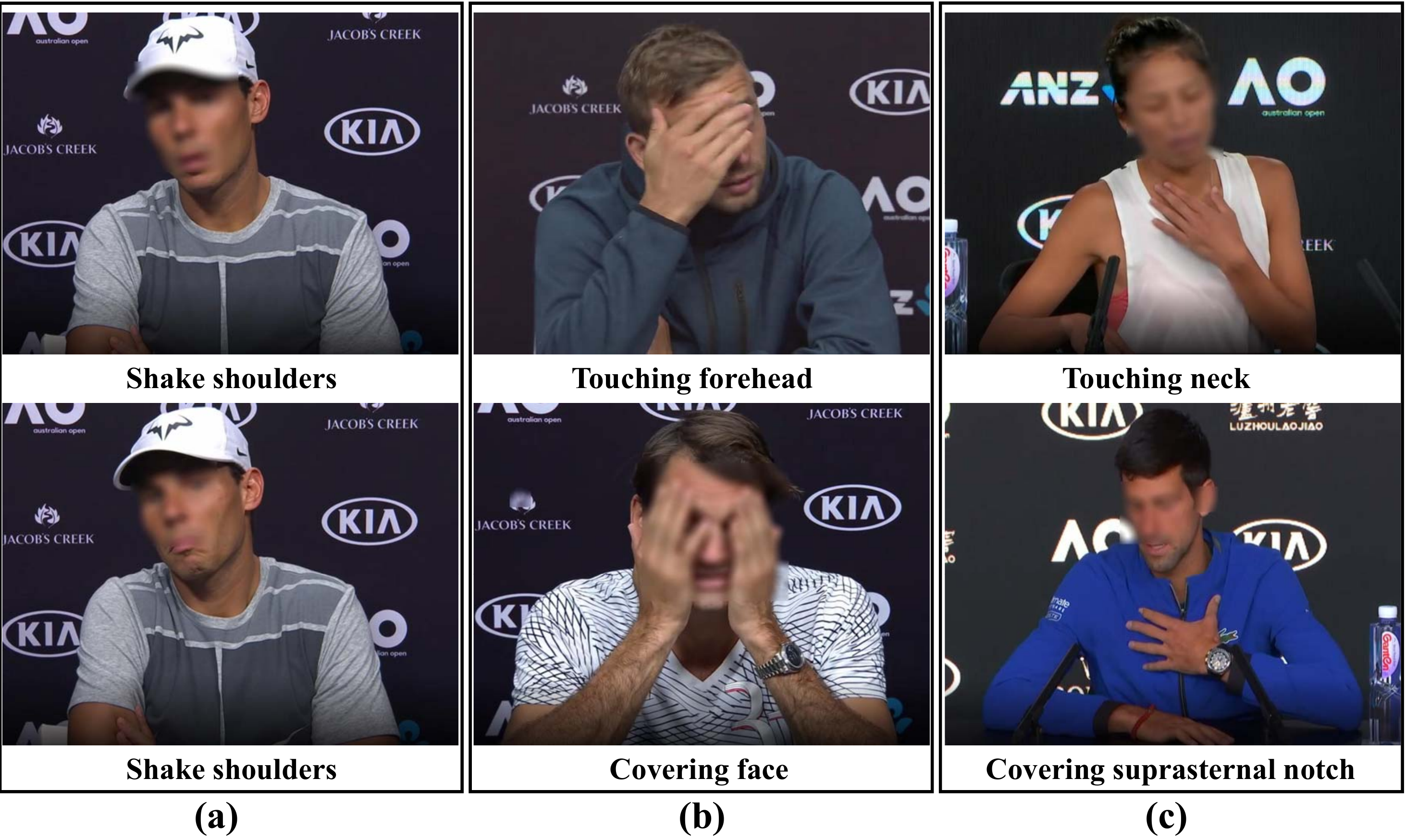}
\end{center}
\vspace{-0.42cm}
\caption{Examples of challenging recognition of the micro-gestures.}
\label{fig:similar}
\end{figure}

From Table \ref{tb:mgreg}, we can summarize several observations: 1) Almost all of methods' accuracy (Top-1) stuck under 60 percentage, which could verify that recognizing such hardly noticeable MGs (\textit{e}.\textit{g}., a short-timing ``Shake shoulders'' as shown in Fig. \ref{fig:similar} (a)) is a very challenging task. Due to the subtle differences between MGs (\textit{e}.\textit{g}., ``Covering face'' vs. ``Touching forehead'', ``Touching neck'' vs. ``Covering suprasternal notch'' as shown in Fig. \ref{fig:similar} (b) and (c)), visual or structural appearances in the form of RGB or pose contribute significantly less than that in a regular gesture (action) recognition task. 2) 3DCNN and RNN-based models' Top-1 performance are lower than 35$\%$, which is not surprising as fully-supervised learning models may have a significant performance drop with class imbalance issue. 3) Capturing temporal dynamics (temporal reasoning) is important as 2DCNN-based TSM and TRN outperform others by large margins. 4) Our method outperforms prior unsupervised learning model P\&C. Although not using any labels, our performance is very competitive with the supervised 3DCNN and RNN-based methods.

\subsection{Video-level Emotion Understanding}
\label{sec:video}
Evaluations also include quantitative analysis comparing performance of the-state-of-art methods for the task of video-level emotion understanding. Here, three RGB-based models with good performance on MG recognition, namely I3D \cite{carreira2017quo}, TRN \cite{zhou2018temporal}, and TSM \cite{lin2019tsm} are selected for comparison. Similarly, two pose-based methods ST-GCN \cite{yan2018spatial} and MS-G3D \cite{liu2020disentangling} are chosen. Those models follow the same configurations with the clip-level recognition. For example, in the task of clip-level MG recognition, TSM \cite{lin2019tsm} divides a clip (input) into 8 segments and samples one RGB frame from each segment to form the input. Now, in the video-level emotion understanding, only the input is changed to a video. We report the performances of these networks in Table \ref{tb:emoreg}, we can see the TSM \cite{lin2019tsm} and TRN \cite{zhou2018temporal} only obtain an emotion classification accuracy of 53 percentage by using the RGB frames as input. The ST-GCN \cite{yan2018spatial} and MS-G3D \cite{liu2020disentangling} yield the similar results with the pose data as input. It is noted that I3D \cite{carreira2017quo} achieves the best score among those models. I3D concatenates outputs from multi-parallel branches at the same level but with different resolutions, which can provide richer representations in emotion classification.

\begin{table}
  \centering
   \footnotesize
    \begin{tabular}{|l|c|c|}
    \hline Methods & Model + Modality & Accuracy\\
    \hline \hline
            TRN \cite{zhou2018temporal}        & \multirow{3}{*}{\tabincell{l}{CNN + RGB}}                & 0.53 		\\
            TSM \cite{lin2019tsm}              &        & 0.53  	\\
            I3D \cite{carreira2017quo}         &                 & \underline{0.57}      \\
    \hline
            ST-GCN \cite{yan2018spatial}       &  \multirow{2}{*}{\tabincell{l}{GCN + Pose}}           & 0.50 \\
            MS-G3D \cite{liu2020disentangling} &          & 0.55 \\
    \hline
            U-S-VAE + LSTM                     & \multirow{2}{*}{\tabincell{l}{RNN + Micro-gesture}}         & 0.55 \\
            TSM \cite{lin2019tsm} + LSTM       &          & \textbf{0.60} \\
     \hline
    \end{tabular}
\caption{Comparison of emotion understanding accuracy (\%) with methods on iMiGUE dataset (best: bold, second best: underlined).}
\vspace{-0.2cm}
\label{tb:emoreg}
\end{table}%

In order to experimentally confirm how micro-gesture influences the classification accuracy of emotions, we feed the probability vectors yield by TSM \cite{lin2019tsm} into a RNN network. This is aiming to train an emotion understanding model via the results of clip-level MG recognition. More specifically, a vector (\textit{e}.\textit{g}., an output of the Softmax layer) that represents the probability distributions of a list of potential outcomes (possible MG classes), will be fed in a three-layered LSTM network (TSM \cite{lin2019tsm} with LSTM). After all vectors (clips) of a video are fed in, the labels of positive or negative (winning or losing of the match) can be used to train the network to understand the emotional states behind a series of micro-gestures in a holistic way. The reason why we select the TSM is because it obtains the best accuracy score in clip-level MG recognition. For comparison, the output vectors of proposed U-S-VAE are also utilized to train a similar emotion understanding network. In Table \ref{tb:emoreg}, we report the emotion understanding results of these models, according to the protocol (video-level) described in Sec. \ref{sec:eva}. We can observe and conclude that micro-gesture is helpful for the emotion understanding. TSM with LSTM (TSM + LSTM) can achieve the best score, U-S-VAE with a 35$\%$ MG recognition accuracy (U-S-VAE + LSTM) can beat most of the compared methods which further verifies that the MG-based analysis can benefit the final emotion understanding.

\subsection{Analysis and Discussion}
To test the generalization capability of U-S-VAE, we provide its performance on different datasets, such as the NTU RGB+D 60 \cite{shahroudy2016ntu}, which is a large scale dataset commonly used for testing action/gesture models. NTU RGB+D contains 60 action categories collected from 40 subjects. In video capturing, each action is recorded simultaneously by three cameras at different horizontal angles. As such, not merely provided the commonly cross-subject (C-Sub) protocol, the authors of NTU RGB+D also recommended the cross-view (C-View) evaluation. We follow this convention and report the recognition accuracy (Top-1) of the two protocols. Here, the results of RGB-D-based unsupervised methods are presented, including Shuffle \& Learn \cite{misra2016shuffle}, and models of Luo \textit{et al.} \cite{luo2017unsupervised} and Li \textit{et al.} \cite{li2018unsupervised}. These models rely on the depth information which are not available in iMiGUE so that we cannot evaluate their performance on our dataset. Moreover, two pose-based models, the LongT GAN \cite{zheng2018unsupervised} and P\&C \cite{su2020predict} are compared for evaluation analysis. Because the code of LongT GAN is not released, we cannot report its result on iMiGUE. Finally, for ablation studies on encoder with different number of BLSTM layer, U-S-VAE with 2-layers BLSTM (U-S-VAE 2L) and 3-layers (U-S-VAE 3L) are chosen for comparison. Ablation studies on different features for classification are carried, and we select the cell states of BLSTM to serve as the feature vectors for testing (U-S-VAE 1L w C). In Table \ref{tb:abl}, we report these experimental results.

\begin{table}[t]
  \centering
   \footnotesize
\begin{tabular}{|l|c|c|c|c|}
    \hline
    \multirow{2}{*}{\tabincell{l}{Unsupervised\\Methods}}  & \multirow{2}{*}{\tabincell{l}{Modality}} & iMiGUE & \multicolumn{2}{c|}{NTU RGB+D}\cr\cline{3-5}
    & & C-Sub & C-View & C-Sub\\
    \hline \hline
            Shuffle \& Learn \cite{misra2016shuffle}            & \multirow{3}{*}{\tabincell{l}{RGB-D}}   & -              & 40.90 & 46.20\\
            Luo \textit{et al.} \cite{luo2017unsupervised}      &    & -              & \underline{53.20}  & \underline{61.40}\\
            Li \textit{et al.} \cite{li2018unsupervised}        &    & -              & \textbf{63.90}     & \textbf{68.10}\\
    \hline
            LongT GAN \cite{zheng2018unsupervised}              & \multirow{6}{*}{\tabincell{l}{Pose}}  & -                  & 48.10 & 39.10\\
            P\&C \cite{su2020predict}                           &   & 31.67              & \textbf{76.10} & \underline{50.70}\\
            U-S-VAE 3L                                          &   & 30.85              & 25.46 & 22.50\\
            U-S-VAE 2L                                          &   & 30.30              & 55.13 & 37.03\\
            U-S-VAE 1L w C                                      &   & \underline{32.04}  & 44.57 & 36.11\\
            U-S-VAE 1L                                          &   & \textbf{32.43}     & \underline{64.88} & \textbf{50.96}\\
     \hline
\end{tabular}
\caption{Ablation study of U-S-VAE with different datasets (best: bold, second best: underlined).}
\vspace{-0.2cm}
\label{tb:abl}
\end{table}%

From the results of Table \ref{tb:abl}, we can summarize several observations: 1) The mutual information (KL-divergence) plays a strong role in learning latent representation since our model (U-S-VAE 1L) can achieve better performance than P\&C \cite{su2020predict} on two datasets with cross-subject protocol. 2) The score of P\&C in cross-view evaluation is higher than ours, this is because P\&C has a pre-processing step to implement a view-invariant transformation \cite{su2020predict}. Besides, P\&C also has an additional feature-level auto-encoder can benefit the classification. 3) In ablation studies, U-S-VAE 1L can achieve the best performance on both iMiGUE and NTU RGB+D datasets. U-S-VAE with hidden states $Z_T$ can beat U-S-VAE with the cell states (U-S-VAE 1L w C).

\vspace{-0.2cm}
\section{Conclusions}
\vspace{-0.1cm}
In this paper, we propose iMiGUE, a new dataset focusing on micro-gestures study. This work not merely investigates representative methods at the MG recognition level, but also attempt to understand the emotional states by using those MGs. We hope these efforts could facilitate new advances in the emotion AI field. In the future, more efforts will be put on studying relationships between MG groups and emotional states. Also, sophisticated models will be explored to find the latent mapping among emotions and MGs in a more holistic way.

{\small
\bibliographystyle{ieee_fullname}
\bibliography{egbib}

\begin{thebibliography}{10}\itemsep=-1pt

\bibitem{aung2015automatic}
Min~SH Aung, Sebastian Kaltwang, Bernardino Romera-Paredes, Brais Martinez,
  Aneesha Singh, Matteo Cella, Michel Valstar, Hongying Meng, Andrew Kemp,
  Moshen Shafizadeh, et~al.
\newblock The automatic detection of chronic pain-related expression:
  requirements, challenges and the multimodal emopain dataset.
\newblock {\em IEEE Trans. Affect. Comput.}, 7(4):435--451, 2015.

\bibitem{aviezer2012body}
Hillel Aviezer, Yaacov Trope, and Alexander Todorov.
\newblock Body cues, not facial expressions, discriminate between intense
  positive and negative emotions.
\newblock {\em Science}, 338(6111):1225--1229, 2012.

\bibitem{axtell1991gestures}
Roger~E Axtell.
\newblock {\em Gestures: the do's and taboos of body language around the wor
  ld}.
\newblock 1991.

\bibitem{bartlett2006fully}
Marian~Stewart Bartlett, Gwen Littlewort, Mark Frank, Claudia Lainscsek, Ian
  Fasel, and Javier Movellan.
\newblock Fully automatic facial action recognition in spontaneous behavior.
\newblock In {\em Proc. IEEE Int. Conf. Auto. Face Gesture Recognit.}, pages
  223--230. IEEE, 2006.

\bibitem{baveye2015liris}
Yoann Baveye, Emmanuel Dellandrea, Christel Chamaret, and Liming Chen.
\newblock {LIRIS-ACCEDE: A} video database for affective content analysis.
\newblock {\em IEEE Trans. Affect. Comput.}, 6(1):43--55, 2015.

\bibitem{burgoon1989nonverbal}
Judee~K Burgoon, David~B Buller, and William~Gill Woodall.
\newblock {\em Nonverbal communication: The unspoken dialogue}.
\newblock Harpercollins College Division, 1989.

\bibitem{cao2017realtime}
Zhe Cao, Tomas Simon, Shih-En Wei, and Yaser Sheikh.
\newblock Realtime multi-person {2D} pose estimation using part affinity
  fields.
\newblock In {\em Proc. IEEE Conf. Comput. Vis. Pattern Recognit.}, pages
  7291--7299, 2017.

\bibitem{carreira2017quo}
Joao Carreira and Andrew Zisserman.
\newblock Quo vadis, action recognition? a new model and the kinetics dataset.
\newblock In {\em Proc. IEEE Conf. Comput. Vis. Pattern Recognit.}, pages
  6299--6308, 2017.

\bibitem{castellano2007recognising}
Ginevra Castellano, Santiago~D Villalba, and Antonio Camurri.
\newblock Recognising human emotions from body movement and gesture dynamics.
\newblock In {\em Int. Conf. Affect. Comput. Intell. Interact.}, pages 71--82.
  Springer, 2007.

\bibitem{cheng2020skeleton}
Ke Cheng, Yifan Zhang, Xiangyu He, Weihan Chen, Jian Cheng, and Hanqing Lu.
\newblock Skeleton-based action recognition with shift graph convolutional
  network.
\newblock In {\em Proc. IEEE Conf. Comput. Vis. Pattern Recognit.}, pages
  183--192, 2020.

\bibitem{davison2016samm}
Adrian~K Davison, Cliff Lansley, Nicholas Costen, Kevin Tan, and Moi~Hoon Yap.
\newblock {SAMM: A} spontaneous micro-facial movement dataset.
\newblock {\em IEEE Trans. Affect. Comput.}, 9(1):116--129, 2016.

\bibitem{dhall2017individual}
Abhinav Dhall, Roland Goecke, Shreya Ghosh, Jyoti Joshi, Jesse Hoey, and Tom
  Gedeon.
\newblock From individual to group-level emotion recognition: {Emotiw 5.0}.
\newblock In {\em Proc. ACM Int. Conf. Multimodal Interaction}, pages 524--528,
  2017.

\bibitem{donahue2015long}
Jeffrey Donahue, Lisa Anne~Hendricks, Sergio Guadarrama, Marcus Rohrbach,
  Subhashini Venugopalan, Kate Saenko, and Trevor Darrell.
\newblock Long-term recurrent convolutional networks for visual recognition and
  description.
\newblock In {\em Proc. IEEE Conf. Comput. Vis. Pattern Recognit.}, pages
  2625--2634, 2015.

\bibitem{douglas2007humaine}
Ellen Douglas-Cowie, Roddy Cowie, Ian Sneddon, Cate Cox, Orla Lowry, Margaret
  Mcrorie, Jean-Claude Martin, Laurence Devillers, Sarkis Abrilian, Anton
  Batliner, et~al.
\newblock The humaine database: Addressing the collection and annotation of
  naturalistic and induced emotional data.
\newblock In {\em Int. Conf. Affect. Comput. Intell. Interact.}, pages
  488--500. Springer, 2007.

\bibitem{du2015hierarchical}
Yong Du, Wei Wang, and Liang Wang.
\newblock Hierarchical recurrent neural network for skeleton based action
  recognition.
\newblock In {\em Proc. IEEE Conf. Comput. Vis. Pattern Recognit.}, pages
  1110--1118, 2015.

\bibitem{ekman2009telling}
Paul Ekman.
\newblock {\em Telling lies: Clues to deceit in the marketplace, politics, and
  marriage (revised edition)}.
\newblock WW Norton \& Company, 2009.

\bibitem{feichtenhofer2019slowfast}
Christoph Feichtenhofer, Haoqi Fan, Jitendra Malik, and Kaiming He.
\newblock Slowfast networks for video recognition.
\newblock In {\em Proc. IEEE Int. Conf. Comput. Vis.}, pages 6202--6211, 2019.

\bibitem{feichtenhofer2016convolutional}
Christoph Feichtenhofer, Axel Pinz, and Andrew Zisserman.
\newblock Convolutional two-stream network fusion for video action recognition.
\newblock In {\em Proc. IEEE Conf. Comput. Vis. Pattern Recognit.}, pages
  1933--1941, 2016.

\bibitem{fourati2014emilya}
Nesrine Fourati and Catherine Pelachaud.
\newblock {Emilya: Emotional} body expression in daily actions database.
\newblock In {\em LREC}, pages 3486--3493, 2014.

\bibitem{gavrilescu2015recognizing}
Mihai Gavrilescu.
\newblock Recognizing emotions from videos by studying facial expressions, body
  postures and hand gestures.
\newblock In {\em 23rd Telecommunications Forum Telfor}, pages 720--723. IEEE,
  2015.

\bibitem{ginger2007gestalt}
Serge Ginger.
\newblock {\em Gestalt therapy: the art of contact}.
\newblock Karnac Books, 2007.

\bibitem{glowinski2008technique}
Donald Glowinski, Antonio Camurri, Gualtiero Volpe, Nele Dael, and Klaus
  Scherer.
\newblock Technique for automatic emotion recognition by body gesture analysis.
\newblock In {\em Proc. IEEE Conf. Comput. Vis. Pattern Recognit. Workshops},
  pages 1--6. IEEE, 2008.

\bibitem{gorban2015thumos}
Alex Gorban, Haroon Idrees, Yu-Gang Jiang, A~Roshan Zamir, Ivan Laptev, Mubarak
  Shah, and Rahul Sukthankar.
\newblock {THUMOS} challenge: Action recognition with a large number of
  classes, 2015.

\bibitem{goyal2017something}
Raghav Goyal, Samira~Ebrahimi Kahou, Vincent Michalski, Joanna Materzynska,
  Susanne Westphal, Heuna Kim, Valentin Haenel, Ingo Fruend, Peter Yianilos,
  Moritz Mueller-Freitag, et~al.
\newblock {The "Something Something"} video database for learning and
  evaluating visual common sense.
\newblock In {\em Proc. IEEE Int. Conf. Comput. Vis.}, volume~1, 2017.

\bibitem{gross2010multi}
Ralph Gross, Iain Matthews, Jeffrey Cohn, Takeo Kanade, and Simon Baker.
\newblock Multi-pie.
\newblock {\em Image Vis. Comput.}, 28(5):807--813, 2010.

\bibitem{gunes2006bimodal}
Hatice Gunes and Massimo Piccardi.
\newblock A bimodal face and body gesture database for automatic analysis of
  human nonverbal affective behavior.
\newblock In {\em Proc. IAPR Int. Conf. Pattern Recognit.}, volume~1, pages
  1148--1153. IEEE, 2006.

\bibitem{hara2018can}
Kensho Hara, Hirokatsu Kataoka, and Yutaka Satoh.
\newblock Can spatiotemporal {3D CNNs} retrace the history of {2D CNNs} and
  imagenet?
\newblock In {\em Proc. IEEE Conf. Comput. Vis. Pattern Recognit.}, pages
  6546--6555, 2018.

\bibitem{joo2015panoptic}
Hanbyul Joo, Hao Liu, Lei Tan, Lin Gui, Bart Nabbe, Iain Matthews, Takeo
  Kanade, Shohei Nobuhara, and Yaser Sheikh.
\newblock Panoptic studio: A massively multiview system for social motion
  capture.
\newblock In {\em Proceedings of the IEEE International Conference on Computer
  Vision}, pages 3334--3342, 2015.

\bibitem{joo2019towards}
Hanbyul Joo, Tomas Simon, Mina Cikara, and Yaser Sheikh.
\newblock Towards social artificial intelligence: Nonverbal social signal
  prediction in a triadic interaction.
\newblock In {\em Proc. IEEE Conf. Comput. Vis. Pattern Recognit.}, pages
  10873--10883, 2019.

\bibitem{kanade2000comprehensive}
Takeo Kanade, Jeffrey~F Cohn, and Yingli Tian.
\newblock Comprehensive database for facial expression analysis.
\newblock In {\em Proc. IEEE Int. Conf. Auto. Face Gesture Recognit.}, pages
  46--53. IEEE, 2000.

\bibitem{kim2019self}
Dahun Kim, Donghyeon Cho, and In~So Kweon.
\newblock Self-supervised video representation learning with space-time cubic
  puzzles.
\newblock In {\em AAAI Conf. Artif. Intell.}, volume~33, pages 8545--8552,
  2019.

\bibitem{kipp2009gesture}
Michael Kipp and Jean-Claude Martin.
\newblock Gesture and emotion: Can basic gestural form features discriminate
  emotions?
\newblock In {\em Int. Conf. Affect. Comput. Intell. Interact. Workshops},
  pages 1--8. IEEE, 2009.

\bibitem{kleinsmith2011automatic}
Andrea Kleinsmith, Nadia Bianchi-Berthouze, and Anthony Steed.
\newblock Automatic recognition of non-acted affective postures.
\newblock {\em IEEE Trans. Syst. Man Cybern. B Cybern.}, 41(4):1027--1038,
  2011.

\bibitem{koelstra2011deap}
Sander Koelstra, Christian Muhl, Mohammad Soleymani, Jong-Seok Lee, Ashkan
  Yazdani, Touradj Ebrahimi, Thierry Pun, Anton Nijholt, and Ioannis Patras.
\newblock Deap: A database for emotion analysis; using physiological signals.
\newblock {\em IEEE Trans. Affect. Comput.}, 3(1):18--31, 2011.

\bibitem{kollias2019deep}
Dimitrios Kollias, Panagiotis Tzirakis, Mihalis~A Nicolaou, Athanasios
  Papaioannou, Guoying Zhao, Bj{\"o}rn Schuller, Irene Kotsia, and Stefanos
  Zafeiriou.
\newblock Deep affect prediction in-the-wild: {Aff-wild} database and
  challenge, deep architectures, and beyond.
\newblock {\em Int. J. Comput. Vision}, 127(6-7):907--929, 2019.

\bibitem{kuehne2011hmdb}
Hildegard Kuehne, Hueihan Jhuang, Est{\'\i}baliz Garrote, Tomaso Poggio, and
  Thomas Serre.
\newblock {HMDB}: a large video database for human motion recognition.
\newblock In {\em Proc. IEEE Int. Conf. Comput. Vis.}, pages 2556--2563. IEEE,
  2011.

\bibitem{lee2017unsupervised}
Hsin-Ying Lee, Jia-Bin Huang, Maneesh Singh, and Ming-Hsuan Yang.
\newblock Unsupervised representation learning by sorting sequences.
\newblock In {\em Proc. IEEE Int. Conf. Comput. Vis.}, pages 667--676, 2017.

\bibitem{li2018spatio}
Chaolong Li, Zhen Cui, Wenming Zheng, Chunyan Xu, and Jian Yang.
\newblock Spatio-temporal graph convolution for skeleton based action
  recognition.
\newblock In {\em AAAI Conf. Artif. Intell.}, 2018.

\bibitem{li2018unsupervised}
Junnan Li, Yongkang Wong, Qi Zhao, and Mohan~S Kankanhalli.
\newblock Unsupervised learning of view-invariant action representations.
\newblock In {\em Proc. Adv. Neural Inf. Process. Syst.}, pages 1254--1264,
  2018.

\bibitem{li2018independently}
Shuai Li, Wanqing Li, Chris Cook, Ce Zhu, and Yanbo Gao.
\newblock Independently recurrent neural network {(IndRNN)}: Building a longer
  and deeper rnn.
\newblock In {\em Proc. IEEE Conf. Comput. Vis. Pattern Recognit.}, pages
  5457--5466, 2018.

\bibitem{li2013spontaneous}
Xiaobai Li, Tomas Pfister, Xiaohua Huang, Guoying Zhao, and Matti
  Pietik{\"a}inen.
\newblock A spontaneous micro-expression database: Inducement, collection and
  baseline.
\newblock In {\em Proc. IEEE Int. Conf. Auto. Face Gesture Recognit.}, pages
  1--6. IEEE, 2013.

\bibitem{li2016online}
Yanghao Li, Cuiling Lan, Junliang Xing, Wenjun Zeng, Chunfeng Yuan, and Jiaying
  Liu.
\newblock Online human action detection using joint classification-regression
  recurrent neural networks.
\newblock In {\em Proc. Eur. Conf. Comput. Vis.}, pages 203--220. Springer,
  2016.

\bibitem{lin2019tsm}
Ji Lin, Chuang Gan, and Song Han.
\newblock {TSM: T}emporal shift module for efficient video understanding.
\newblock In {\em Proc. IEEE Int. Conf. Comput. Vis.}, pages 7083--7093, 2019.

\bibitem{liu2016spatio}
Jun Liu, Amir Shahroudy, Dong Xu, and Gang Wang.
\newblock Spatio-temporal {LSTM} with trust gates for {3D} human action
  recognition.
\newblock In {\em Proc. Eur. Conf. Comput. Vis.}, pages 816--833. Springer,
  2016.

\bibitem{liu2017global}
Jun Liu, Gang Wang, Ping Hu, Ling-Yu Duan, and Alex~C Kot.
\newblock Global context-aware attention {LSTM} networks for {3D} action
  recognition.
\newblock In {\em Proc. IEEE Conf. Comput. Vis. Pattern Recognit.}, pages
  1647--1656, 2017.

\bibitem{liu2020disentangling}
Ziyu Liu, Hongwen Zhang, Zhenghao Chen, Zhiyong Wang, and Wanli Ouyang.
\newblock Disentangling and unifying graph convolutions for skeleton-based
  action recognition.
\newblock In {\em Proc. IEEE Conf. Comput. Vis. Pattern Recognit.}, pages
  143--152, 2020.

\bibitem{lucey2011painful}
Patrick Lucey, Jeffrey~F Cohn, Kenneth~M Prkachin, Patricia~E Solomon, and Iain
  Matthews.
\newblock Painful data: The unbc-mcmaster shoulder pain expression archive
  database.
\newblock In {\em Proc. IEEE Int. Conf. Auto. Face Gesture Recognit.}, pages
  57--64. IEEE, 2011.

\bibitem{luo2020arbee}
Yu Luo, Jianbo Ye, Reginald~B Adams, Jia Li, Michelle~G Newman, and James~Z
  Wang.
\newblock {ARBEE: T}owards automated recognition of bodily expression of
  emotion in the wild.
\newblock {\em Int. J. Comput. Vision}, 128(1):1--25, 2020.

\bibitem{luo2017unsupervised}
Zelun Luo, Boya Peng, De-An Huang, Alexandre Alahi, and Li Fei-Fei.
\newblock Unsupervised learning of long-term motion dynamics for videos.
\newblock In {\em Proc. IEEE Conf. Comput. Vis. Pattern Recognit.}, pages
  2203--2212, 2017.

\bibitem{Mahasseni_regularized_lstm:2016}
B. Mahasseni and S. Todorovic.
\newblock Regularizing long short term memory with {3D} human-skeleton
  sequences for action recognition.
\newblock In {\em Proc. IEEE Conf. Comput. Vis. Pattern Recognit.}, pages
  3054--3062. IEEE, June 2016.

\bibitem{mckeown2011semaine}
Gary McKeown, Michel Valstar, Roddy Cowie, Maja Pantic, and Marc Schroder.
\newblock The semaine database: Annotated multimodal records of emotionally
  colored conversations between a person and a limited agent.
\newblock {\em IEEE Trans. Affect. Comput.}, 3(1):5--17, 2011.

\bibitem{misra2016shuffle}
Ishan Misra, C~Lawrence Zitnick, and Martial Hebert.
\newblock Shuffle and learn: unsupervised learning using temporal order
  verification.
\newblock In {\em Proc. Eur. Conf. Comput. Vis.}, pages 527--544. Springer,
  2016.

\bibitem{monfort2019moments}
Mathew Monfort, Alex Andonian, Bolei Zhou, Kandan Ramakrishnan, Sarah~Adel
  Bargal, Tom Yan, Lisa Brown, Quanfu Fan, Dan Gutfreund, Carl Vondrick, et~al.
\newblock Moments in time dataset: one million videos for event understanding.
\newblock {\em IEEE Trans. Pattern Anal. Mach. Intell.}, 42(2):502--508, 2019.

\bibitem{navarro2016every}
Joe Navarro and Marvin Karlins.
\newblock {\em What every body is saying}.
\newblock HarperCollins, 2016.

\bibitem{noroozi2018survey}
Fatemeh Noroozi, Dorota Kaminska, Ciprian Corneanu, Tomasz Sapinski, Sergio
  Escalera, and Gholamreza Anbarjafari.
\newblock Survey on emotional body gesture recognition.
\newblock {\em IEEE Trans. Affect. Comput.}, 2018.

\bibitem{pantic2005web}
Maja Pantic, Michel Valstar, Ron Rademaker, and Ludo Maat.
\newblock Web-based database for facial expression analysis.
\newblock In {\em Proc. IEEE Int. Conf. Multimed. Expo.}, pages 317--321. IEEE,
  2005.

\bibitem{pease2008definitive}
Barbara Pease and Allan Pease.
\newblock {\em The definitive book of body language: The hidden meaning behind
  people's gestures and expressions}.
\newblock Bantam, 2008.

\bibitem{peng2020learning}
Wei Peng, Xiaopeng Hong, Haoyu Chen, and Guoying Zhao.
\newblock Learning graph convolutional network for skeleton-based human action
  recognition by neural searching.
\newblock In {\em AAAI Conf. Artif. Intell.}, pages 2669--2676, 2020.

\bibitem{psaltis2016multimodal}
Athanasios Psaltis, Kyriaki Kaza, Kiriakos Stefanidis, Spyridon Thermos,
  Konstantinos~C Apostolakis, Kosmas Dimitropoulos, and Petros Daras.
\newblock Multimodal affective state recognition in serious games applications.
\newblock In {\em IEEE Int. Conf. Imaging Sys. Tech.}, pages 435--439. IEEE,
  2016.

\bibitem{ranganathan2016multimodal}
Hiranmayi Ranganathan, Shayok Chakraborty, and Sethuraman Panchanathan.
\newblock Multimodal emotion recognition using deep learning architectures.
\newblock In {\em IEEE Winter Conf. Appl. Comput. Vis.}, pages 1--9. IEEE,
  2016.

\bibitem{ringeval2013introducing}
Fabien Ringeval, Andreas Sonderegger, Juergen Sauer, and Denis Lalanne.
\newblock Introducing the {RECOLA} multimodal corpus of remote collaborative
  and affective interactions.
\newblock In {\em Proc. IEEE Int. Conf. Auto. Face Gesture Recognit.}, pages
  1--8. IEEE, 2013.

\bibitem{rohrbach2016recognizing}
Marcus Rohrbach, Anna Rohrbach, Michaela Regneri, Sikandar Amin, Mykhaylo
  Andriluka, Manfred Pinkal, and Bernt Schiele.
\newblock Recognizing fine-grained and composite activities using hand-centric
  features and script data.
\newblock {\em Int. J. Comput. Vision}, 119(3):346--373, 2016.

\bibitem{saha2014study}
Sriparna Saha, Shreyasi Datta, Amit Konar, and Ramadoss Janarthanan.
\newblock A study on emotion recognition from body gestures using {Kinect}
  sensor.
\newblock In {\em Int. Conf. Signal Process. Commun.}, pages 056--060. IEEE,
  2014.

\bibitem{schindler2008recognizing}
Konrad Schindler, Luc Van~Gool, and Beatrice De~Gelder.
\newblock Recognizing emotions expressed by body pose: {A} biologically
  inspired neural model.
\newblock {\em Neural Networks}, 21(9):1238--1246, 2008.

\bibitem{schuller2011avec}
Bj{\"o}rn Schuller, Michel Valstar, Florian Eyben, Gary McKeown, Roddy Cowie,
  and Maja Pantic.
\newblock Avec 2011--the first international audio/visual emotion challenge.
\newblock In {\em Int. Conf. Affect. Comput. Intell. Interact.}, pages
  415--424. Springer, 2011.

\bibitem{schuller2012avec}
Bj{\"o}rn Schuller, Michel Valster, Florian Eyben, Roddy Cowie, and Maja
  Pantic.
\newblock Avec 2012: the continuous audio/visual emotion challenge.
\newblock In {\em Proc. ACM Int. Conf. Multimodal Interact.}, pages 449--456,
  2012.

\bibitem{shahroudy2016ntu}
Amir Shahroudy, Jun Liu, Tian-Tsong Ng, and Gang Wang.
\newblock {NTU RGB+D: A} large scale dataset for {3D} human activity analysis.
\newblock In {\em Proc. IEEE Conf. Comput. Vis. Pattern Recognit.}, pages
  1010--1019, 2016.

\bibitem{shi2018bidirectional}
Henglin Shi, Xin Liu, Xiaopeng Hong, and Guoying Zhao.
\newblock Bidirectional long short-term memory variational autoencoder.
\newblock In {\em BMVC}, page 165, 2018.

\bibitem{shi2019two}
Lei Shi, Yifan Zhang, Jian Cheng, and Hanqing Lu.
\newblock Two-stream adaptive graph convolutional networks for skeleton-based
  action recognition.
\newblock In {\em Proc. IEEE Conf. Comput. Vis. Pattern Recognit.}, pages
  12026--12035, 2019.

\bibitem{si2019attention}
Chenyang Si, Wentao Chen, Wei Wang, Liang Wang, and Tieniu Tan.
\newblock An attention enhanced graph convolutional {LSTM} network for
  skeleton-based action recognition.
\newblock In {\em Proc. IEEE Conf. Comput. Vis. Pattern Recognit.}, pages
  1227--1236, 2019.

\bibitem{sigurdsson2016hollywood}
Gunnar~A Sigurdsson, G{\"u}l Varol, Xiaolong Wang, Ali Farhadi, Ivan Laptev,
  and Abhinav Gupta.
\newblock Hollywood in homes: Crowdsourcing data collection for activity
  understanding.
\newblock In {\em Proc. Eur. Conf. Comput. Vis.}, pages 510--526. Springer,
  2016.

\bibitem{simonyan2014two}
Karen Simonyan and Andrew Zisserman.
\newblock Two-stream convolutional networks for action recognition in videos.
\newblock In {\em Proc. Adv. Neural Inf. Process. Syst.}, pages 568--576, 2014.

\bibitem{soleymani2011multimodal}
Mohammad Soleymani, Jeroen Lichtenauer, Thierry Pun, and Maja Pantic.
\newblock A multimodal database for affect recognition and implicit tagging.
\newblock {\em IEEE Trans. Affect. Comput.}, 3(1):42--55, 2011.

\bibitem{song2017end}
Sijie Song, Cuiling Lan, Junliang Xing, Wenjun Zeng, and Jiaying Liu.
\newblock An end-to-end spatio-temporal attention model for human action
  recognition from skeleton data.
\newblock In {\em Proc. AAAI Conf. Artif. Intell.}, pages 4263--4270, 2017.

\bibitem{soomro2012ucf101}
Khurram Soomro, Amir~Roshan Zamir, and Mubarak Shah.
\newblock Ucf101: A dataset of 101 human actions classes from videos in the
  wild.
\newblock {\em CRCV-TR-12-01}, 2012.

\bibitem{srivastava2015unsupervised}
Nitish Srivastava, Elman Mansimov, and Ruslan Salakhudinov.
\newblock Unsupervised learning of video representations using {LSTMs}.
\newblock In {\em Int. Conf. Mach. Learn.}, pages 843--852, 2015.

\bibitem{su2020predict}
Kun Su, Xiulong Liu, and Eli Shlizerman.
\newblock {Predict \& Cluster: Unsupervised} skeleton based action recognition.
\newblock In {\em Proc. IEEE Conf. Comput. Vis. Pattern Recognit.}, pages
  9631--9640, 2020.

\bibitem{tao2020self}
Li Tao, Xueting Wang, and Toshihiko Yamasaki.
\newblock Self-supervised video representation learning using inter-intra
  contrastive framework.
\newblock In {\em Proc. ACM Int. Conf. Multimed.}, pages 2193--2201, 2020.

\bibitem{tran2015learning}
Du Tran, Lubomir Bourdev, Rob Fergus, Lorenzo Torresani, and Manohar Paluri.
\newblock Learning spatiotemporal features with {3D} convolutional networks.
\newblock In {\em Proc. IEEE Int. Conf. Comput. Vis.}, pages 4489--4497, 2015.

\bibitem{tran2019video}
Du Tran, Heng Wang, Lorenzo Torresani, and Matt Feiszli.
\newblock Video classification with channel-separated convolutional networks.
\newblock In {\em Proc. IEEE Int. Conf. Comput. Vis.}, pages 5552--5561, 2019.

\bibitem{tran2018closer}
Du Tran, Heng Wang, Lorenzo Torresani, Jamie Ray, Yann LeCun, and Manohar
  Paluri.
\newblock A closer look at spatiotemporal convolutions for action recognition.
\newblock In {\em Proc. IEEE Conf. Comput. Vis. Pattern Recognit.}, pages
  6450--6459, 2018.

\bibitem{valstar2010induced}
Michel Valstar and Maja Pantic.
\newblock Induced disgust, happiness and surprise: an addition to the mmi
  facial expression database.
\newblock In {\em Proc. Int. Conf. Language Resources and Evaluation, Workshop
  EMOTION}, pages 65--70. Paris, France., 2010.

\bibitem{veeriah2015differential}
Vivek Veeriah, Naifan Zhuang, and Guo-Jun Qi.
\newblock Differential recurrent neural networks for action recognition.
\newblock In {\em Proc. IEEE Int. Conf. Comput. Vis.}, pages 4041--4049, 2015.

\bibitem{vinciarelli2009social}
Alessandro Vinciarelli, Maja Pantic, and Herv{\'e} Bourlard.
\newblock Social signal processing: Survey of an emerging domain.
\newblock {\em Image Vis. Comput.}, 27(12):1743--1759, 2009.

\bibitem{wang2019self}
Jiangliu Wang, Jianbo Jiao, Linchao Bao, Shengfeng He, Yunhui Liu, and Wei Liu.
\newblock Self-supervised spatio-temporal representation learning for videos by
  predicting motion and appearance statistics.
\newblock In {\em Proc. IEEE Conf. Comput. Vis. Pattern Recognit.}, pages
  4006--4015, 2019.

\bibitem{wang2018temporal}
Limin Wang, Yuanjun Xiong, Zhe Wang, Yu Qiao, Dahua Lin, Xiaoou Tang, and Luc
  Van~Gool.
\newblock Temporal segment networks for action recognition in videos.
\newblock {\em IEEE Trans. Pattern Anal. Mach. Intell.}, 41(11):2740--2755,
  2018.

\bibitem{wang2018non}
Xiaolong Wang, Ross Girshick, Abhinav Gupta, and Kaiming He.
\newblock Non-local neural networks.
\newblock In {\em Proc. IEEE Conf. Comput. Vis. Pattern Recognit.}, pages
  7794--7803, 2018.

\bibitem{xie2018rethinking}
Saining Xie, Chen Sun, Jonathan Huang, Zhuowen Tu, and Kevin Murphy.
\newblock Rethinking spatiotemporal feature learning: {Speed-accuracy}
  trade-offs in video classification.
\newblock In {\em Proc. Eur. Conf. Comput. Vis.}, pages 305--321, 2018.

\bibitem{xu2019self}
Dejing Xu, Jun Xiao, Zhou Zhao, Jian Shao, Di Xie, and Yueting Zhuang.
\newblock Self-supervised spatiotemporal learning via video clip order
  prediction.
\newblock In {\em Proc. IEEE Conf. Comput. Vis. Pattern Recognit.}, pages
  10334--10343, 2019.

\bibitem{yan2018spatial}
Sijie Yan, Yuanjun Xiong, and Dahua Lin.
\newblock Spatial temporal graph convolutional networks for skeleton-based
  action recognition.
\newblock In {\em AAAI Conf. Artif. Intell.}, 2018.

\bibitem{yan2014casme}
Wen-Jing Yan, Xiaobai Li, Su-Jing Wang, Guoying Zhao, Yong-Jin Liu, Yu-Hsin
  Chen, and Xiaolan Fu.
\newblock Casme ii: An improved spontaneous micro-expression database and the
  baseline evaluation.
\newblock {\em PloS one}, 9(1):e86041, 2014.

\bibitem{yin20063d}
Lijun Yin, Xiaozhou Wei, Yi Sun, Jun Wang, and Matthew~J Rosato.
\newblock A 3d facial expression database for facial behavior research.
\newblock In {\em Proc. IEEE Int. Conf. Auto. Face Gesture Recognit.}, pages
  211--216. IEEE, 2006.

\bibitem{yu2020humbi}
Zhixuan Yu, Jae~Shin Yoon, In~Kyu Lee, Prashanth Venkatesh, Jaesik Park, Jihun
  Yu, and Hyun~Soo Park.
\newblock {HUMBI: A Large Multiview Dataset of Human Body Expressions}.
\newblock In {\em Proc. IEEE Conf. Comput. Vis. Pattern Recognit.}, pages
  2990--3000, 2020.

\bibitem{yu2020searching}
Zitong Yu, Benjia Zhou, Jun Wan, Pichao Wang, Haoyu Chen, Xin Liu, Stan~Z Li,
  and Guoying Zhao.
\newblock Searching multi-rate and multi-modal temporal enhanced networks for
  gesture recognition.
\newblock {\em arXiv preprint arXiv:2008.09412}, 2020.

\bibitem{zhang2017view}
Pengfei Zhang, Cuiling Lan, Junliang Xing, Wenjun Zeng, Jianru Xue, and Nanning
  Zheng.
\newblock View adaptive recurrent neural networks for high performance human
  action recognition from skeleton data.
\newblock In {\em Proc. IEEE Int. Conf. Comput. Vis.}, pages 2117--2126, 2017.

\bibitem{zhang2013high}
Xing Zhang, Lijun Yin, Jeffrey~F Cohn, Shaun Canavan, Michael Reale, Andy
  Horowitz, and Peng Liu.
\newblock A high-resolution spontaneous 3d dynamic facial expression database.
\newblock In {\em Proc. IEEE Int. Conf. Auto. Face Gesture Recognit.}, pages
  1--6. IEEE, 2013.

\bibitem{zheng2018unsupervised}
Nenggan Zheng, Jun Wen, Risheng Liu, Liangqu Long, Jianhua Dai, and Zhefeng
  Gong.
\newblock Unsupervised representation learning with long-term dynamics for
  skeleton based action recognition.
\newblock In {\em AAAI Conf. Artif. Intell.}, 2018.

\bibitem{zhou2018temporal}
Bolei Zhou, Alex Andonian, Aude Oliva, and Antonio Torralba.
\newblock Temporal relational reasoning in videos.
\newblock In {\em Proc. Eur. Conf. Comput. Vis.}, pages 803--818, 2018.

\bibitem{zhu2016co}
Wentao Zhu, Cuiling Lan, Junliang Xing, Wenjun Zeng, Yanghao Li, Li Shen,
  Xiaohui Xie, et~al.
\newblock Co-occurrence feature learning for skeleton based action recognition
  using regularized deep {LSTM} networks.
\newblock In {\em Proc. AAAI Conf. Artif. Intell.}, volume~2, page~8, 2016.

\end{thebibliography}
}

\end{document}